\documentclass[11pt, a4paper, logo, copyright, nonumbering]{xiaomi}

\usepackage[numbers, sort&compress, square]{natbib}
\usepackage{dblfloatfix}
\usepackage[normalem]{ulem}
\usepackage{caption}
\usepackage{dramatist}
\usepackage{xspace}
\usepackage{pifont}
\usepackage{multirow}
\usepackage{tcolorbox}
\tcbuselibrary{breakable,listings,skins}
\usepackage{fancyvrb}
\usepackage{xltabular}
\usepackage{longtable}
\usepackage{hyperref}
\usepackage{wrapfig}
\usepackage{placeins}

\usepackage{amsfonts}
\usepackage{amsmath}
\usepackage{amssymb}
\usepackage{lineno}
\usepackage{adjustbox}

\usepackage[bottom]{footmisc}

\usepackage{CJKutf8}
\usepackage{setspace}
\usepackage{makecell}
\usepackage{graphicx}
\usepackage{subcaption}
\usepackage{multicol}

\usepackage[utf8]{inputenc}
\usepackage[T1]{fontenc}
\usepackage[varqu]{zi4} 
\usepackage{cleveref}
\usepackage{url}
\usepackage{booktabs}
\usepackage{nicefrac}
\usepackage{microtype}
\usepackage{xcolor}
\usepackage{colortbl}

\definecolor{BrickRed}{rgb}{.72,0,0}
\definecolor{darkgreen}{rgb}{0.0, 0.5, 0.0}
\definecolor{ForestGreen}{RGB}{34,139,34}
\definecolor{LakeBlue}{RGB}{0,61,153}
\definecolor{MiOrange}{RGB}{255,225,204}
\definecolor{Hex}{RGB}{225,213,231}

\AtBeginDocument{%
  \renewcommand{\bfseries}{\usefont{T1}{XCharter-TLF}{b}{n}}%
}

\hypersetup{
    colorlinks=true,
    allcolors=LakeBlue
}

\title{\centering STAR: SpatioTemporal Adaptive Reward Allocation for Text-to-Image RL Post-Training}
\titlerunning{STAR}

\author{
  Jinjie Shen, Wei Deng, Xian Hu, Daiguo Zhou, Jian Luan
}
\institute{
  {\normalfont\fontsize{10.5}{12}\selectfont Correspondence to shenjinjie22@gmail.com}
}

\begin{document}

\begin{abstract}
Existing RL post-training methods for text-to-image generation usually convert the final-image reward into a single scalar advantage and apply it with the same strength to the entire generative trajectory. However, text-to-image generation naturally has temporal and spatial structure: different denoising steps are responsible for different generation stages, and the content that truly determines text alignment often appears only in part of the image. This granularity mismatch makes it difficult for policy updates to focus on the generative components that actually affect the reward. To address this issue, we propose \textbf{SpatioTemporal Adaptive Reward (STAR) Allocation} for RL post-training of text-to-image diffusion and flow models. STAR uses text-image attention inside the generative model and starts from the core content that the user truly cares about in the prompt. It constructs spatial allocation maps that dynamically vary across denoising steps and rollouts, and allocates the same group-relative advantage to more relevant latent regions with almost no additional computational overhead. STAR then applies stronger policy updates to these regions through a spatially resolved policy objective. We use Stable Diffusion 3.5 Medium as the base model and evaluate on three tasks: GenEval, OCR text rendering, and PickScore. Experimental results show that STAR improves compositional semantic alignment, text rendering, and preference optimization without changing the external reward source, achieving $\mathbf{0.9759}$, $\mathbf{0.9757}$, and $\mathbf{23.60}$ on GenEval, OCR, and PickScore, respectively.
\end{abstract}

\maketitle

\begin{figure*}[t]
  \centering
  \includegraphics[width=\textwidth]{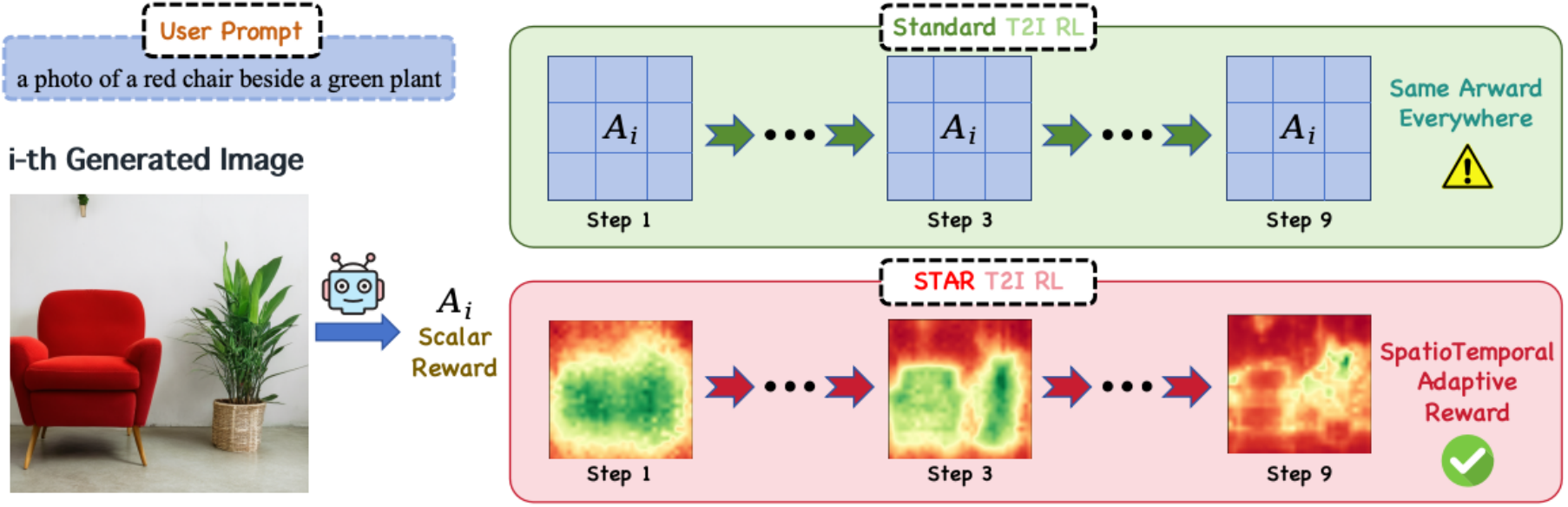}
  \caption{Overview of STAR (SpatioTemporal Adaptive Reward Allocation). Given the same image-level group-relative advantage, STAR extracts text-image attention maps from the generative backbone and constructs timestep-specific spatial allocation maps that highlight latent regions corresponding to key prompt components. These maps route the scalar advantage into spatially resolved training signals, so policy updates selectively focus on the generative regions that determine text alignment.}
  \label{fig:workflow}
\end{figure*}

\section{Introduction}

Diffusion models have become mainstream frameworks for high-fidelity text-to-image generation~\cite{ho2020denoising,dhariwal2021diffusion,song2020score,ddim,rombach2022high}. Recent text-to-image systems further improve resolution and language understanding through latent diffusion, rectified-flow transformers, and large-scale prompt-image training~\cite{podell2023sdxl,drawbench,sd3}. Although these models have made substantial progress in visual quality, compositional generalization under complex prompts remains a major weakness: models often make errors in object counting, attribute binding, and spatial relations. Since such errors can usually only be judged on the completed image by automatic evaluators, preference models~\cite{xu2024imagereward,kirstain2023pick}, or human feedback, RL-style post-training provides an effective path for directly optimizing image-level objectives.

A key limitation of existing RL post-training for text-to-image generation lies in the granularity of credit assignment. The final reward is usually an image-level scalar, whereas the diffusion/flow generative trajectory has explicit temporal and spatial structure: sampling spans multiple denoising/flow steps, and each step contains many latent spatial tokens. However, an image-level reward does not mean that the entire image bears equal responsibility for the reward. Even for a simple prompt such as ``a bicycle,'' text alignment is mainly determined by the region containing the bicycle itself, rather than by background regions such as the sky or the road. When the prompt further contains colors, counts, and spatial relations, the reward depends more strongly on whether a small number of semantically relevant regions are generated correctly. This issue is particularly visible in online diffusion/flow RL post-training methods, including Flow-GRPO~\cite{liu2025flowgrpo}, where a group-relative reward is still converted into a scalar advantage for policy optimization. Existing policy-gradient training usually applies this advantage with the same strength to all timesteps and spatial locations, making it difficult to distinguish the reward responsibility of subject regions from that of background regions.

This mismatch stems from the spatiotemporal structure of diffusion and flow generative trajectories~\cite{lipman2022flow,rectified_flow}. Different denoising stages affect global layout, object formation, and detail refinement, respectively; within each stage, only a subset of latent regions is directly related to the subjects, attributes, and relations in the current prompt. We study spatiotemporal credit assignment in this structured process: given the same image-level outcome feedback, we allocate it over denoising time $\times$ latent space so that policy updates focus on the generative components that truly determine text alignment.

We propose \textbf{SpatioTemporal Adaptive Reward (STAR) Allocation}, a framework for attention-based T2I diffusion and flow post-training. STAR uses text-image interactions inside the generative model to construct timestep-specific routing maps and converts them into latent-space allocation maps. Given the same scalar advantage, STAR distributes its training strength across denoising steps and latent regions. Combined with a spatially resolved likelihood-ratio objective, regions corresponding to key prompt components receive stronger updates, while background and weakly related regions maintain weaker influence.

We instantiate STAR in compositional T2I RL post-training and compare it with scalar-allocation baselines under the same external reward source. Experimental results show that STAR achieves $0.9759$, $23.60$, and $0.9757$ on GenEval, PickScore, and OCR, respectively.

Our contributions are as follows:
\begin{itemize}
    \item We propose \textbf{SpatioTemporal Adaptive Reward (STAR) Allocation}, a method for T2I diffusion/flow RL post-training. STAR converts image-level rewards into training signals over denoising time and latent space, allowing updates to focus on the generative regions corresponding to key objects, attributes, and relations in the prompt.
    \item We use text-image interactions inside the model to construct timestep-specific spatial allocation maps, extend existing RL post-training objectives into a spatially resolved policy objective, and provide a unified reward-routing interface for attention-based T2I diffusion/flow models.
    \item We validate STAR on GenEval, PickScore, and OCR, where it achieves $0.9759$, $23.60$, and $0.9757$, respectively, showing that SpatioTemporal Adaptive Reward (STAR) Allocation can simultaneously improve compositional semantic alignment, preference optimization, and text rendering.
\end{itemize}

\section{Related Work}

\textbf{RL post-training for T2I.}
In recent years, directly optimizing downstream rewards has become an important direction for improving diffusion and flow-based T2I models, following the broader policy-gradient view of reinforcement learning~\cite{sutton1998reinforcement,williams1992simple}. DDPO~\cite{ddpo} formulates the denoising process as a multi-step decision-making problem and optimizes diffusion models with policy-gradient methods. DPOK~\cite{fan2024reinforcement} further introduces online RL and KL regularization into T2I diffusion fine-tuning. Diffusion-DPO~\cite{wallace2024diffusion} builds on direct preference optimization~\cite{dpo}, while AlignProp~\cite{prabhudesai2023aligning} optimizes human preferences through differentiable reward backpropagation. More recently, Flow-GRPO~\cite{liu2025flowgrpo} extends online policy-gradient training to flow matching models and enables efficient post-training through stochastic sampling and denoising reduction. These methods demonstrate the effectiveness of RL-style post-training for T2I models, but they usually treat the final reward as a scalar signal applied to the whole image or the entire trajectory. STAR studies how the same reward signal should be allocated within a structured denoising trajectory.

\textbf{Fine-grained rewards and credit assignment.}
Existing work has begun to recognize that image-level scalar rewards provide overly coarse feedback for diffusion alignment. A Dense Reward View~\cite{yang2024dense} revisits preference alignment from a sequential decision-making perspective and derives a dense objective that emphasizes early denoising steps. B\textsuperscript{2}-DiffuRL addresses sparse terminal rewards by estimating the contributions of intermediate denoising actions through backward progressive training and branch-based sampling. Another line of work improves the reward source itself. ImageReward~\cite{xu2024imagereward} learns a general T2I human preference reward model; ImageDoctor further provides multi-aspect scores and grounded heatmaps for diagnosing local errors in generated images. SpatialReward builds verifiable reward modeling for spatial consistency. These methods highlight the value of fine-grained feedback in T2I post-training, while leaving a complementary direction: given an existing outcome reward, how can we use the generative model's own text-image interactions to allocate this signal within the spatiotemporal structure?

\section{Method}

STAR reallocates an existing image-level reward to the structured generative process of diffusion/flow models. Given a prompt $c$, the model generates a latent trajectory
\begin{equation}
\tau=(z_T,z_{T-1},\ldots,z_0), \qquad
z_t\in\mathbb{R}^{C\times h_t\times w_t},
\end{equation}
where $z_0$ corresponds to the final image. Let $\Omega_t=\{1,\ldots,h_t\}\times\{1,\ldots,w_t\}$ denote the set of spatial locations in the latent feature map at step $t$. The external reward source provides scalar feedback only on the final sample, denoted as
\begin{equation}
R_i=R(z_0^i,c), \qquad A_i=\mathcal{B}(R_i,c),
\end{equation}
where $i$ denotes the $i$-th rollout under the same prompt, and $A_i$ is the sample-level advantage obtained by group-relative normalization or baseline subtraction. A standard RL-style objective uses $A_i$ as a trajectory-level weight, as in group-relative policy optimization~\cite{grpo}; STAR routes $A_i$ to denoising steps and latent regions.

\subsection{Attention-Guided Allocation}

STAR represents each prompt as a set of reward-relevant text units:
\begin{equation}
\mathcal{K}(c)=\{q_1,\ldots,q_K\},
\end{equation}
where $q_k$ denotes a core component in the prompt. These components are the conditions that the final image needs to satisfy, including subjects, styles, attributes, counts, and spatial relations. For example, for the prompt ``generate a futuristic-style building,'' $\mathcal{K}(c)$ contains the subject ``building'' and the visual style requirement ``futuristic style.'' Dataset-specific extraction rules and prompt templates for text-unit extraction are provided in Appendix~\ref{app:text_unit_prompts}.

Suppose the full prompt $c$ is converted by the text encoder tokenizer into $L_c$ tokens. Each text unit $q_k$ corresponds to several text tokens, whose token indices are denoted as
\begin{equation}
\mathcal{T}_k \subseteq \{1,\ldots,L_c\}.
\end{equation}

STAR reads routing signals from a selected set of text-image attention layers $\mathcal{L}$. When generating rollout $i$, attention-based T2I backbones establish interactions between image latent tokens and text tokens at each denoising step $t$. For any $\ell\in\mathcal{L}$, this layer provides an attention map $\alpha_t^{i,\ell}\in\mathbb{R}^{h_t\times w_t\times L_c}$; the attention weight from latent position $u$ to text token $j$ is denoted as
\begin{equation}
\alpha_{t}^{i,\ell}(u,j).
\end{equation}
Here $u\in\Omega_t$ is a latent spatial position and $j$ is a text token index. In joint-attention backbones (e.g., Stable Diffusion 3.5, Qwen-Image), this value corresponds to the image-to-text block of the full attention matrix; in cross-attention backbones (e.g., Stable Diffusion XL), it corresponds to the cross-attention weights from image queries to text keys. STAR takes the maximum image-to-text attention weight over the selected layers:
\begin{equation}
\alpha_t^i(u,j)=\max_{\ell\in\mathcal{L}}\alpha_{t}^{i,\ell}(u,j),
\end{equation}
which gives the timestep-specific text-image routing map $\alpha_t^i\in\mathbb{R}^{h_t\times w_t\times L_c}$ for rollout $i$ at step $t$.

Next, STAR extracts where each text unit appears in the current trajectory. Given the token set $\mathcal{T}_k$ of text unit $q_k$, STAR constructs its spatial map $M_t^{i,k}\in\mathbb{R}^{h_t\times w_t}$, where the routing strength at latent position $u$ is
\begin{equation}
M_t^{i,k}(u)=\frac{1}{|\mathcal{T}_k|}\sum_{j\in\mathcal{T}_k}\alpha_t^i(u,j).
\end{equation}
STAR aggregates these key components into a trajectory-timestep specific heatmap $H_t^i\in\mathbb{R}^{h_t\times w_t}$:
\begin{equation}
H_t^i(u)=\frac{1}{K}\sum_{k=1}^{K} M_t^{i,k}(u).
\end{equation}
This heatmap specifies where the prompt-relevant content is most active. To turn it into a normalized reward-allocation map, STAR maps it to bounded coefficients $C_t^i\in\mathbb{R}^{h_t\times w_t}$:
\begin{equation}
C_t^i(u)=1+(\lambda-1)
\frac{H_t^i(u)-\min_{v\in\Omega_t}H_t^i(v)}
{\max_{v\in\Omega_t}H_t^i(v)-\min_{v\in\Omega_t}H_t^i(v)+\epsilon}.
\end{equation}
Here $\lambda\ge1$ controls the allocation strength, and $\epsilon$ ensures numerical stability. Larger $C_t^i(u)$ strengthens updates on prompt-relevant regions; smaller but nonzero coefficients preserve the influence of background, composition, and overall quality on the final reward.

\subsection{Spatially Resolved Policy Objective}

STAR uses the allocation map to convert the sample-level advantage into a spatialized training signal. For the $i$-th rollout and denoising step $t$, we define the spatial advantage map as
\begin{equation}
A_t^i(u)=A_i\,C_t^i(u), \qquad A_t^i\in\mathbb{R}^{h_t\times w_t}.
\end{equation}
This mapping preserves the sign of $A_i$ and increases the update magnitude for prompt-relevant regions through $C_t^i(u)$.

The policy update acts on spatial log-probability contributions. Let
\begin{equation}
\ell_{\theta,t}^i\in\mathbb{R}^{h_t\times w_t}, \qquad
\bar{\ell}_{\theta,t}^i=\frac{1}{|\Omega_t|}\sum_{u\in\Omega_t}\ell_{\theta,t}^i(u),
\end{equation}
where $\ell_{\theta,t}^i(u)$ denotes the log-probability contribution at latent position $u$, and $\bar{\ell}_{\theta,t}^i$ corresponds to the scalar transition log-probability used in standard policy-gradient training. STAR uses a normalized spatial ratio:
\begin{equation}
\rho_{\theta,t}^i(u)=
\operatorname{sg}\!\left[\exp\!\left(\bar{\ell}_{\theta,t}^i-\bar{\ell}_{\mathrm{old},t}^i\right)\right]
\frac{\exp(\ell_{\theta,t}^i(u))}{\operatorname{sg}\!\left[\exp(\ell_{\theta,t}^i(u))\right]},
\end{equation}
where $\bar{\ell}_{\mathrm{old},t}^i$ is the rollout log-probability stored during sampling, and $\operatorname{sg}[\cdot]$ denotes stop-gradient. This ratio preserves the overall scale of the scalar likelihood ratio while assigning gradients over spatial locations.

Finally, STAR optimizes the clipped spatial objective, following the clipped-ratio policy update used in PPO-style training~\cite{ppo}:
\begin{equation}
\mathcal{J}_{\mathrm{STAR}}(\theta)=
\mathbb{E}\left[
\frac{1}{T}\sum_{t=1}^{T}\frac{1}{|\Omega_t|}\sum_{u\in\Omega_t}
\min\left(
 \rho_{\theta,t}^i(u)A_t^i(u),
 \operatorname{clip}(\rho_{\theta,t}^i(u),1-\varepsilon,1+\varepsilon)A_t^i(u)
\right)
\right]-\beta\mathcal{D}_{\mathrm{KL}}.
\end{equation}
Here $\mathcal{D}_{\mathrm{KL}}$ is the KL regularization term between the current policy and the frozen reference policy over denoising transitions, which constrains the update magnitude and reduces the risk of reward hacking.

\section{Experiments}

\subsection{Implementation Details}

\textbf{Evaluation tasks.}
The experiments include three T2I post-training tasks: GenEval, OCR text rendering, and PickScore. GenEval evaluates compositional generation ability, including object counting, attribute binding, colors, and spatial relations~\cite{geneval,T2i-compbench}; training prompts are generated from the official GenEval templates, and rewards are computed by the official rule-based evaluation pipeline. OCR text rendering requires the model to generate a specified text string in the image, and the reward is computed from the edit distance between the rendered text and the target text. PickScore uses a learned preference reward model to score each prompt-image pair.

\textbf{Baselines.}
We compare against multiple representative T2I baselines, including diffusion models (LDM, SD1.5, SD2.1, SD-XL~\cite{podell2023sdxl}, DALLE-2~\cite{ramesh2022hierarchical}, DALLE-3~\cite{betker2023improving}), autoregressive models (Show-o~\cite{xie2024show}, Emu3-Gen~\cite{wang2024emu3}, JanusFlow~\cite{ma2024janusflow}, Janus-Pro-7B~\cite{chen2025janus}), and recent diffusion/flow models (FLUX.1 Dev~\cite{flux2024}, SD3.5-L, SANA-1.5 4.8B~\cite{xie2025sana}, SD3.5-M~\cite{sd3}, SD3.5-M+Flow-GRPO~\cite{liu2025flowgrpo}).

\textbf{Experimental setup.}
We implement STAR based on the Flow-GRPO~\cite{liu2025flowgrpo} training pipeline and use Stable Diffusion 3.5 Medium (SD3.5-M) as the base model. Training uses ODE-to-SDE rollouts for online sampling, with the number of sampling steps set to $T=10$, the SDE noise level set to $a=0.7$, and the image resolution set to $512\times512$; evaluation uses $T=40$ steps. For each prompt, we sample $G=24$ images to form a group, and update the model with a group-relative policy-gradient objective, clipped ratio, and KL regularization. The KL coefficient is set to $\beta=0.04$ for GenEval and OCR text rendering, and $\beta=0.01$ for PickScore. We train the model with LoRA~\cite{hu2021lora}, using rank $r=32$ and scaling factor $\alpha=64$. STAR records text-image attention during rollout and uses the spatial objective in Section 3 to allocate group-relative advantages. Attention routing uses the joint attention of SD3.5-M, reads the attention layers from transformer blocks 10--15, and sets the allocation strength to $\lambda=1.5$.

\subsection{Main Results}

\begin{table*}[t]
\centering
\caption{GenEval results. We report the overall score and six compositional sub-tasks. Obj. denotes object, and Attr. denotes attribute. Blue indicates the best result, and green indicates the second-best result.}
\label{tab:geneval}
\setlength{\tabcolsep}{4.5pt}
\renewcommand{\arraystretch}{1.10}
\resizebox{\textwidth}{!}{
\begin{tabular}{lccccccc}
\toprule
\rowcolor{Hex!35}
\textbf{Model} & \textbf{Overall} & \textbf{Single Obj.} & \textbf{Two Obj.} & \textbf{Counting} & \textbf{Colors} & \textbf{Position} & \textbf{Attr. Binding} \\
\midrule
\rowcolor{gray!12}
\multicolumn{8}{c}{\textcolor{gray!70!black}{\textit{Diffusion Models}}} \\
LDM & 0.37 & 0.92 & 0.29 & 0.23 & 0.70 & 0.02 & 0.05 \\
SD1.5 & 0.43 & 0.97 & 0.38 & 0.35 & 0.76 & 0.04 & 0.06 \\
SD2.1 & 0.50 & 0.98 & 0.51 & 0.44 & 0.85 & 0.07 & 0.17 \\
SD-XL & 0.55 & 0.98 & 0.74 & 0.39 & 0.85 & 0.15 & 0.23 \\
DALLE-2 & 0.52 & 0.94 & 0.66 & 0.49 & 0.77 & 0.10 & 0.19 \\
DALLE-3 & 0.67 & 0.96 & 0.87 & 0.47 & 0.83 & 0.43 & 0.45 \\
\midrule
\rowcolor{gray!12}
\multicolumn{8}{c}{\textcolor{gray!70!black}{\textit{Autoregressive Models}}} \\
Show-o & 0.53 & 0.95 & 0.52 & 0.49 & 0.82 & 0.11 & 0.28 \\
Emu3-Gen & 0.54 & 0.98 & 0.71 & 0.34 & 0.81 & 0.17 & 0.21 \\
JanusFlow & 0.63 & 0.97 & 0.59 & 0.45 & 0.83 & 0.53 & 0.42 \\
Janus-Pro-7B & 0.80 & 0.99 & 0.89 & 0.59 & 0.90 & 0.79 & 0.66 \\
\midrule
\rowcolor{gray!12}
\multicolumn{8}{c}{\textcolor{gray!70!black}{\textit{Diffusion / Flow Models}}} \\
FLUX.1 Dev & 0.66 & 0.98 & 0.81 & 0.74 & 0.79 & 0.22 & 0.45 \\
SD3.5-L & 0.71 & 0.98 & 0.89 & 0.73 & 0.83 & 0.34 & 0.47 \\
SANA-1.5 4.8B & 0.81 & 0.99 & 0.93 & 0.86 & 0.84 & 0.59 & 0.65 \\
SD3.5-M & 0.63 & 0.98 & 0.78 & 0.50 & 0.81 & 0.24 & 0.52 \\
SD3.5-M+Flow-GRPO & \textcolor{ForestGreen}{0.95} & \textcolor{LakeBlue}{\textbf{1.00}} & \textcolor{ForestGreen}{0.99} & \textcolor{ForestGreen}{0.95} & \textcolor{ForestGreen}{0.92} & \textcolor{LakeBlue}{\textbf{0.99}} & \textcolor{ForestGreen}{0.86} \\
\midrule
\rowcolor{LakeBlue!8}
\textcolor{LakeBlue}{\textbf{SD3.5-M+Flow-GRPO+STAR}} & \textcolor{LakeBlue}{\textbf{0.9759}} & \textcolor{ForestGreen}{\textbf{1.00}} & \textcolor{LakeBlue}{\textbf{1.00}} & \textcolor{LakeBlue}{\textbf{0.98}} & \textcolor{LakeBlue}{\textbf{0.96}} & \textcolor{ForestGreen}{\textbf{0.98}} & \textcolor{LakeBlue}{\textbf{0.94}} \\
\bottomrule
\end{tabular}
}
\end{table*}

\textbf{GenEval results.}
Table~\ref{tab:geneval} presents the main results on GenEval~\cite{geneval}. STAR achieves an overall score of $0.9759$, outperforming SD3.5-M at $0.63$ and Flow-GRPO~\cite{liu2025flowgrpo} at $0.95$. This improvement mainly comes from more fine-grained compositional semantic constraints: on counting, colors, and attribute binding, STAR reaches $0.98$, $0.96$, and $0.94$, respectively, clearly improving over $0.95$, $0.92$, and $0.86$. Two-object generation further reaches $1.00$. On single-object generation and position, two sub-tasks that are already close to saturation, STAR still maintains near-optimal performance, reaching $1.00$ and $0.98$, respectively.

These results show that the gains of STAR come from more accurate spatiotemporal credit assignment. By allocating the same image-level advantage to prompt-relevant latent regions, STAR can more selectively update the generative regions that affect object counts, colors, and attribute binding, thereby improving compositional generation performance.

\begin{figure}[t]
\centering
\includegraphics[width=0.92\textwidth]{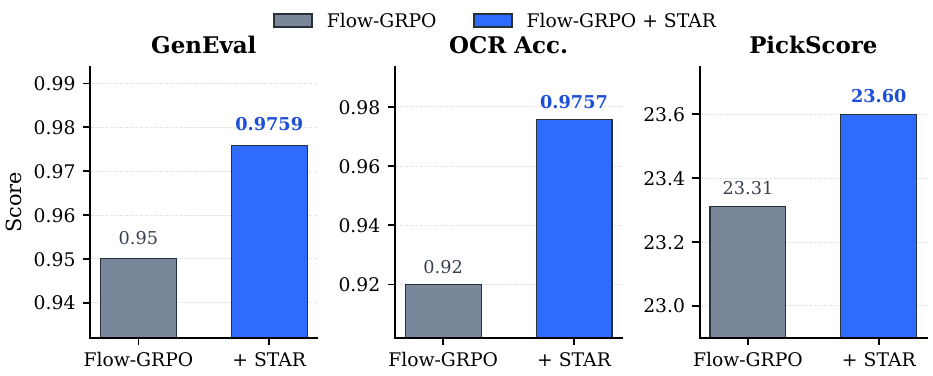}
\caption{Main results on three T2I post-training benchmarks. STAR consistently improves over the Flow-GRPO baseline on compositional generation, text rendering, and preference optimization.}
\label{fig:main_results_three_tasks}
\end{figure}

\textbf{Results across benchmarks.}
Figure~\ref{fig:main_results_three_tasks} further compares Flow-GRPO~\cite{liu2025flowgrpo} and STAR on the three post-training tasks. STAR achieves $0.9759$ on GenEval~\cite{geneval}, $0.9757$ on OCR text rendering~\cite{textdiffuser}, and $23.60$ on PickScore~\cite{kirstain2023pick}, all outperforming the corresponding Flow-GRPO baselines. The improvements on GenEval and OCR indicate that spatiotemporal adaptive reward allocation can more effectively optimize verifiable text-alignment objectives; the improvement on PickScore shows that this allocation scheme can also be used with continuous rewards provided by a preference model. Overall, without changing the form of the task reward, STAR allocates the same sample-level feedback to more relevant generative regions and thus obtains consistent gains in compositional generation, text rendering, and preference optimization.

\subsection{Ablation Study}

\begin{figure}[t]
\centering
\begin{subfigure}[t]{0.46\textwidth}
\centering
\includegraphics[width=\linewidth]{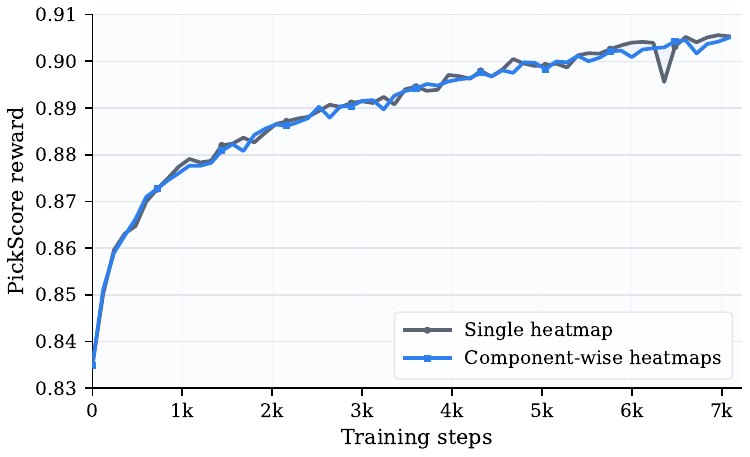}
\caption{Prompt heatmap construction.}
\label{fig:ablation_heatmap_construction}
\end{subfigure}\hfill
\begin{subfigure}[t]{0.46\textwidth}
\centering
\includegraphics[width=\linewidth]{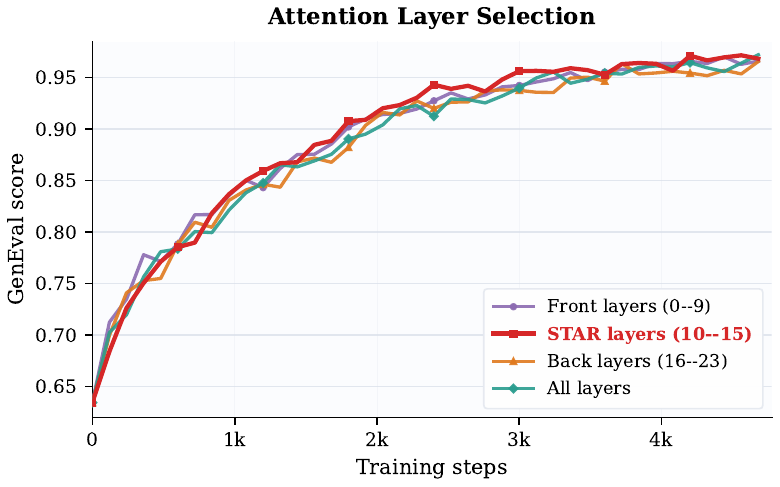}
\caption{Attention layer selection.}
\label{fig:ablation_attention_layers}
\end{subfigure}

\vspace{0.2em}
\begin{subfigure}[t]{0.40\textwidth}
\centering
\includegraphics[width=\linewidth]{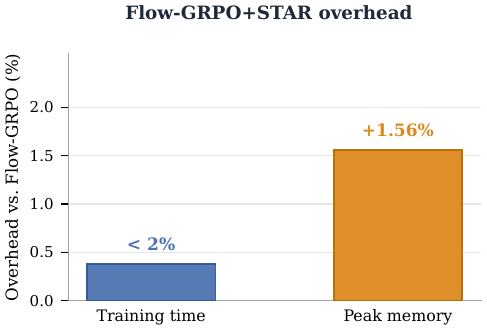}
\caption{Computational overhead.}
\label{fig:compute_overhead_star_vs_scalar}
\end{subfigure}
\caption{Ablation and efficiency analysis for STAR, covering reward routing design choices and additional training overhead.}
\label{fig:star_ablation_summary}
\end{figure}

\subsubsection{Prompt Heatmap Construction}
We compare two ways of constructing prompt heatmaps. The single-heatmap variant first aggregates the attention of multiple key components in the prompt into one heatmap and then obtains reward allocation coefficients; the component-wise heatmaps variant first generates a heatmap for each key component separately and then stacks and normalizes them. Both experiments use the same PickScore reward, training setting, and evaluation interval.

Figure~\ref{fig:ablation_heatmap_construction} shows that the training curves of the two variants are overall very close, with the single heatmap showing a slight advantage in the later stage of training. This result indicates that directly aggregating multiple components that the user cares about in the prompt into a shared heatmap already provides a stable routing signal. Generating heatmaps component by component and then stacking them does not bring additional gains and increases implementation complexity; therefore, we use single-heatmap construction in the main experiments.

\subsubsection{Attention Layer Selection}
We further analyze the choice of text-image attention layers in STAR. We compare four settings: front layers use transformer blocks $0$--$9$, STAR layers use blocks $10$--$15$, back layers use blocks $16$--$23$, and all layers use all attention layers.

Figure~\ref{fig:ablation_attention_layers} shows that layer selection affects the stability of attention routing. Front layers and back layers both bring effective improvements, but their later-stage curves are slightly lower; all layers reach the highest level in the final stage, but require reading more attention maps. STAR layers remain in the best-performing group for most of training and are overall close to all layers, while using only a small number of middle blocks. This indicates that blocks $10$--$15$ already contain sufficiently clear text-image alignment signals. Therefore, we use this layer set in the main experiments to achieve a better balance between effectiveness and implementation overhead.

\subsubsection{Computational Overhead}
We further compare the training overhead of Flow-GRPO~\cite{liu2025flowgrpo} and Flow-GRPO+STAR. STAR reuses the text-image attention from the forward pass of the generative model and only performs heatmap aggregation and advantage rescaling on the selected layers; it does not require training an additional reward model or adding new image evaluation steps.

Figure~\ref{fig:compute_overhead_star_vs_scalar} shows that STAR introduces less than $2\%$ additional training time overhead, and increases peak memory by only $1.56\%$. This result shows that the main computation of STAR still comes from the original diffusion/flow rollout and GRPO update; attention-based allocation is only a lightweight reward-routing step. Therefore, we can obtain more fine-grained spatiotemporal credit assignment with almost no change in training cost.

\subsection{Qualitative Comparison}

To qualitatively examine the effect of post-training with STAR, we compare the original SD3.5-M base model with SD3.5-M after Flow-GRPO+STAR post-training on a diverse set of prompts.

\begin{figure}[t]
\centering
\begin{subfigure}[t]{0.48\textwidth}
\centering
\includegraphics[width=\linewidth]{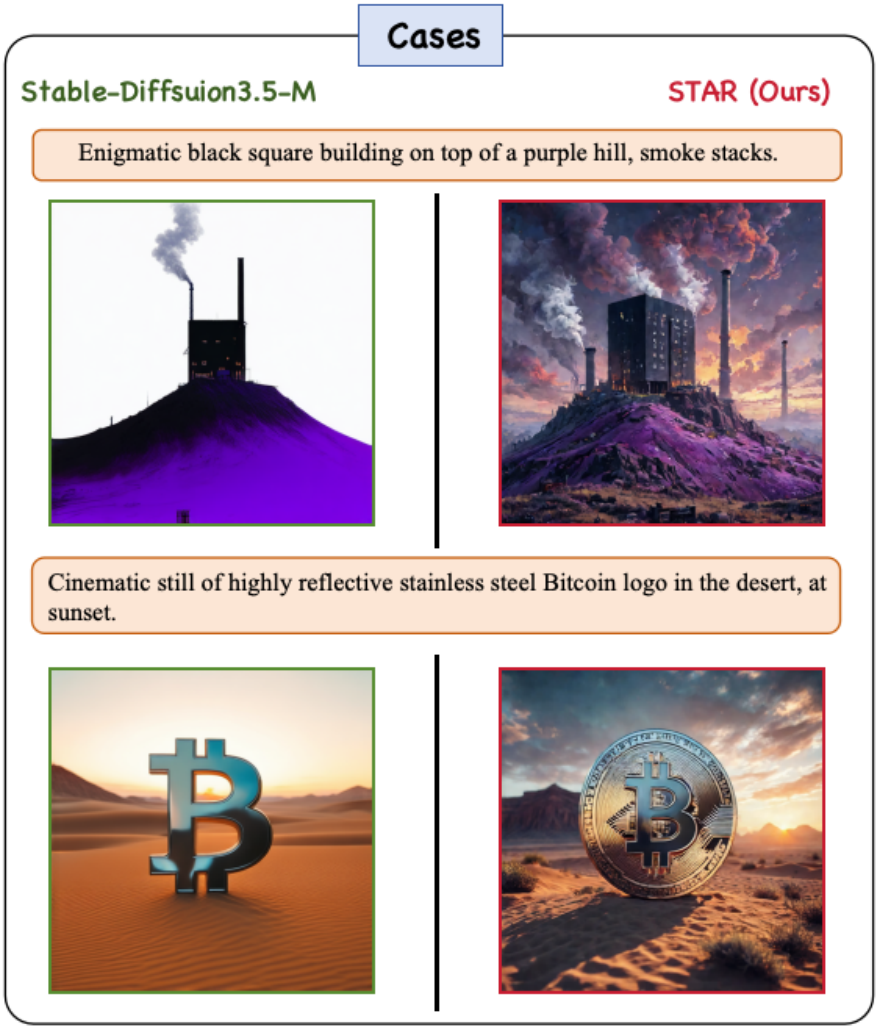}
\end{subfigure}\hfill
\begin{subfigure}[t]{0.48\textwidth}
\centering
\includegraphics[width=\linewidth]{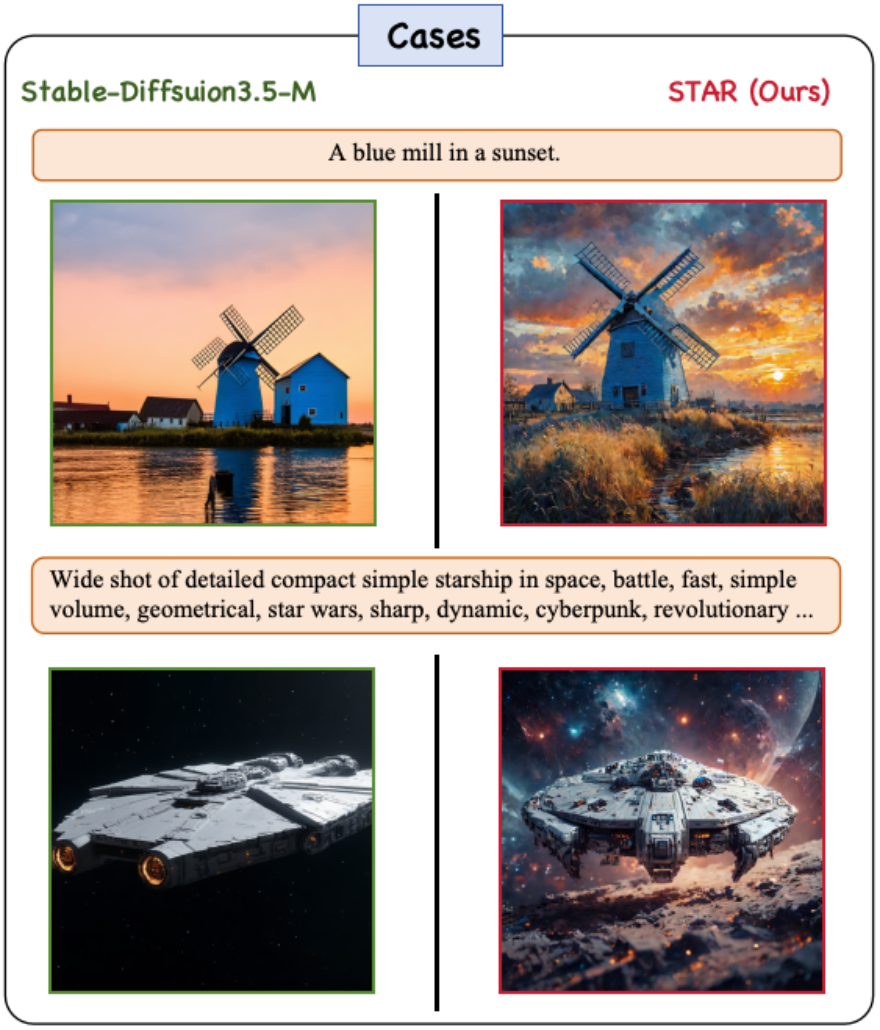}
\end{subfigure}
\caption{Qualitative comparison between the original SD3.5-M and SD3.5-M after Flow-GRPO+STAR post-training. Within every prompt pair, the left image is produced by SD3.5-M and the right image by SD3.5-M+Flow-GRPO+STAR.}
\label{fig:case_study}
\end{figure}

As shown in Figure~\ref{fig:case_study}, both models generate images that capture the main concepts in the prompts. Across the displayed examples, Flow-GRPO+STAR post-training changes how the requested subjects, attributes, and surrounding scenes are realized. For example, the post-trained model depicts the black square building, reflective stainless-steel Bitcoin logo, blue windmill, and compact starship with more complete prompt-related elements and more coherent scene composition.

These qualitative results complement the quantitative evaluations and illustrate the effect of applying Flow-GRPO+STAR post-training to SD3.5-M across different types of prompts. By allocating the reward signal to prompt-relevant latent regions, STAR encourages the model to better incorporate the textual conditions during generation.

\FloatBarrier
\section{Conclusion}

We propose \textbf{SpatioTemporal Adaptive Reward (STAR) Allocation}, a spatiotemporal adaptive reward allocation method for RL post-training of T2I diffusion and flow models. STAR extracts prompt-related spatiotemporal signals from text-image interactions inside the generative model and uses them to allocate image-level feedback more selectively to key regions of the generative trajectory, focusing policy updates on the parts that truly affect text alignment and generation quality. Experimental results show that, without changing the external reward source and with almost no additional training cost, STAR stably improves existing RL post-training methods, especially in scenarios requiring fine-grained semantic understanding, local content control, and preference alignment. Overall, STAR provides a simple, lightweight, and effective interface for allocating image-level feedback to structured diffusion/flow trajectories.

\section*{Statements}
Stable Diffusion 3.5 Medium is used under the Stability AI Community License for academic research purposes. The authors confirm that the use of this model is limited to academic research purposes and has not been used for any commercial activities.

\bibliography{ref}

@inproceedings{liu2025flowgrpo,
  title={Flow-{GRPO}: Training Flow Matching Models via Online {RL}},
  author={Liu, Jie and Liu, Gongye and Liang, Jiajun and Li, Yangguang and Liu, Jiaheng and Wang, Xintao and Wan, Pengfei and Zhang, Di and Ouyang, Wanli},
  booktitle={Advances in Neural Information Processing Systems},
  year={2025}
}

@inproceedings{rombach2022high,
  title={High-resolution image synthesis with latent diffusion models},
  author={Rombach, Robin and Blattmann, Andreas and Lorenz, Dominik and Esser, Patrick and Ommer, Bj{\"o}rn},
  booktitle={Proceedings of the IEEE/CVF conference on computer vision and pattern recognition},
  pages={10684--10695},
  year={2022}
}

@inproceedings{lipman2022flow,
  author       = {Yaron Lipman and
                  Ricky T. Q. Chen and
                  Heli Ben{-}Hamu and
                  Maximilian Nickel and
                  Matthew Le},
  title        = {Flow Matching for Generative Modeling},
  booktitle    = {The Eleventh International Conference on Learning Representations,
                  {ICLR} 2023, Kigali, Rwanda, May 1-5, 2023},
  publisher    = {OpenReview.net},
  year         = {2023},
}

@inproceedings{rectified_flow,
  author       = {Xingchao Liu and
                  Chengyue Gong and
                  Qiang Liu},
  title        = {Flow Straight and Fast: Learning to Generate and Transfer Data with
                  Rectified Flow},
  booktitle    = {The Eleventh International Conference on Learning Representations,
                  {ICLR} 2023, Kigali, Rwanda, May 1-5, 2023},
  year         = {2023},
}

@inproceedings{sd3,
  title={Scaling rectified flow transformers for high-resolution image synthesis},
  author={Esser, Patrick and Kulal, Sumith and Blattmann, Andreas and Entezari, Rahim and M{\"u}ller, Jonas and Saini, Harry and Levi, Yam and Lorenz, Dominik and Sauer, Axel and Boesel, Frederic and others},
  booktitle={Forty-first international conference on machine learning},
  year={2024}
}

@misc{flux2024,
    author={Black Forest Labs},
    title={FLUX},
    year={2024},
    howpublished={\url{https://github.com/black-forest-labs/flux}},
}

@inproceedings{ddpo,
  author       = {Kevin Black and
                  Michael Janner and
                  Yilun Du and
                  Ilya Kostrikov and
                  Sergey Levine},
  title        = {Training Diffusion Models with Reinforcement Learning},
  booktitle    = {The Twelfth International Conference on Learning Representations,
                  {ICLR} 2024, Vienna, Austria, May 7-11, 2024},
  year         = {2024},
}

@article{fan2024reinforcement,
  title={Reinforcement learning for fine-tuning text-to-image diffusion models},
  author={Fan, Ying and Watkins, Olivia and Du, Yuqing and Liu, Hao and Ryu, Moonkyung and Boutilier, Craig and Abbeel, Pieter and Ghavamzadeh, Mohammad and Lee, Kangwook and Lee, Kimin},
  journal={Advances in Neural Information Processing Systems},
  volume={36},
  year={2024}
}

@inproceedings{wallace2024diffusion,
  title={Diffusion model alignment using direct preference optimization},
  author={Wallace, Bram and Dang, Meihua and Rafailov, Rafael and Zhou, Linqi and Lou, Aaron and Purushwalkam, Senthil and Ermon, Stefano and Xiong, Caiming and Joty, Shafiq and Naik, Nikhil},
  booktitle={Proceedings of the IEEE/CVF Conference on Computer Vision and Pattern Recognition},
  pages={8228--8238},
  year={2024}
}

@article{prabhudesai2023aligning,
  title={Aligning text-to-image diffusion models with reward backpropagation},
  author={Prabhudesai, Mihir and Goyal, Anirudh and Pathak, Deepak and Fragkiadaki, Katerina},
  journal={arXiv preprint arXiv:2310.03739},
  year={2023}
}

@inproceedings{yang2024dense,
  author       = {Shentao Yang and
                  Tianqi Chen and
                  Mingyuan Zhou},
  title        = {A Dense Reward View on Aligning Text-to-Image Diffusion with Preference},
  booktitle    = {Forty-first International Conference on Machine Learning, {ICML} 2024,
                  Vienna, Austria, July 21-27, 2024},
  pages        = {55998--56032},
  year         = {2024},
}

@article{xu2024imagereward,
  title={Imagereward: Learning and evaluating human preferences for text-to-image generation},
  author={Xu, Jiazheng and Liu, Xiao and Wu, Yuchen and Tong, Yuxuan and Li, Qinkai and Ding, Ming and Tang, Jie and Dong, Yuxiao},
  journal={Advances in Neural Information Processing Systems},
  volume={36},
  year={2024}
}

@article{geneval,
  title={Geneval: An object-focused framework for evaluating text-to-image alignment},
  author={Ghosh, Dhruba and Hajishirzi, Hannaneh and Schmidt, Ludwig},
  journal={Advances in Neural Information Processing Systems},
  volume={36},
  pages={52132--52152},
  year={2023}
}

@article{textdiffuser,
  title={Textdiffuser: Diffusion models as text painters},
  author={Chen, Jingye and Huang, Yupan and Lv, Tengchao and Cui, Lei and Chen, Qifeng and Wei, Furu},
  journal={Advances in Neural Information Processing Systems},
  volume={36},
  pages={9353--9387},
  year={2023}
}

@article{kirstain2023pick,
  title={Pick-a-pic: An open dataset of user preferences for text-to-image generation},
  author={Kirstain, Yuval and Polyak, Adam and Singer, Uriel and Matiana, Shahbuland and Penna, Joe and Levy, Omer},
  journal={Advances in Neural Information Processing Systems},
  volume={36},
  pages={36652--36663},
  year={2023}
}

@article{T2i-compbench,
  title={T2i-compbench: A comprehensive benchmark for open-world compositional text-to-image generation},
  author={Huang, Kaiyi and Sun, Kaiyue and Xie, Enze and Li, Zhenguo and Liu, Xihui},
  journal={Advances in Neural Information Processing Systems},
  volume={36},
  pages={78723--78747},
  year={2023}
}

@article{grpo,
  title={Deepseekmath: Pushing the limits of mathematical reasoning in open language models},
  author={Shao, Zhihong and Wang, Peiyi and Zhu, Qihao and Xu, Runxin and Song, Junxiao and Bi, Xiao and Zhang, Haowei and Zhang, Mingchuan and Li, YK and Wu, Y and others},
  journal={arXiv preprint arXiv:2402.03300},
  year={2024}
}

@article{ppo,
  title={Proximal policy optimization algorithms},
  author={Schulman, John and Wolski, Filip and Dhariwal, Prafulla and Radford, Alec and Klimov, Oleg},
  journal={arXiv preprint arXiv:1707.06347},
  year={2017}
}

@inproceedings{ddim,
  author       = {Jiaming Song and
                  Chenlin Meng and
                  Stefano Ermon},
  title        = {Denoising Diffusion Implicit Models},
  booktitle    = {9th International Conference on Learning Representations, {ICLR} 2021,
                  Virtual Event, Austria, May 3-7, 2021},
  year         = {2021},
}

@inproceedings{song2020score,
  author       = {Yang Song and
                  Jascha Sohl{-}Dickstein and
                  Diederik P. Kingma and
                  Abhishek Kumar and
                  Stefano Ermon and
                  Ben Poole},
  title        = {Score-Based Generative Modeling through Stochastic Differential Equations},
  booktitle    = {9th International Conference on Learning Representations, {ICLR} 2021,
                  Virtual Event, Austria, May 3-7, 2021},
  year         = {2021},
}

@article{ho2020denoising,
  title={Denoising diffusion probabilistic models},
  author={Ho, Jonathan and Jain, Ajay and Abbeel, Pieter},
  journal={Advances in neural information processing systems},
  volume={33},
  pages={6840--6851},
  year={2020}
}

@article{dhariwal2021diffusion,
  title={Diffusion models beat gans on image synthesis},
  author={Dhariwal, Prafulla and Nichol, Alexander},
  journal={Advances in neural information processing systems},
  volume={34},
  pages={8780--8794},
  year={2021}
}

@inproceedings{podell2023sdxl,
  author       = {Dustin Podell and
                  Zion English and
                  Kyle Lacey and
                  Andreas Blattmann and
                  Tim Dockhorn and
                  Jonas M{\"{u}}ller and
                  Joe Penna and
                  Robin Rombach},
  title        = {{SDXL:} Improving Latent Diffusion Models for High-Resolution Image
                  Synthesis},
  booktitle    = {The Twelfth International Conference on Learning Representations,
                  {ICLR} 2024, Vienna, Austria, May 7-11, 2024},
  year         = {2024},
}

@article{ramesh2022hierarchical,
  title={Hierarchical text-conditional image generation with clip latents},
  author={Ramesh, Aditya and Dhariwal, Prafulla and Nichol, Alex and Chu, Casey and Chen, Mark},
  journal={arXiv preprint arXiv:2204.06125},
  volume={1},
  number={2},
  pages={3},
  year={2022}
}

@article{betker2023improving,
  title={Improving image generation with better captions},
  author={Betker, James and Goh, Gabriel and Jing, Li and Brooks, Tim and Wang, Jianfeng and Li, Linjie and Ouyang, Long and Zhuang, Juntang and Lee, Joyce and Guo, Yufei and others},
  journal={Computer Science. https://cdn. openai. com/papers/dall-e-3. pdf},
  volume={2},
  number={3},
  pages={8},
  year={2023}
}

@article{drawbench,
  title={Photorealistic text-to-image diffusion models with deep language understanding},
  author={Saharia, Chitwan and Chan, William and Saxena, Saurabh and Li, Lala and Whang, Jay and Denton, Emily L and Ghasemipour, Kamyar and Gontijo Lopes, Raphael and Karagol Ayan, Burcu and Salimans, Tim and others},
  journal={Advances in neural information processing systems},
  volume={35},
  pages={36479--36494},
  year={2022}
}

@book{sutton1998reinforcement,
  title={Reinforcement learning: An introduction},
  author={Sutton, Richard S and Barto, Andrew G and others},
  volume={1},
  year={1998},
  publisher={MIT press Cambridge}
}

@article{williams1992simple,
  title={Simple statistical gradient-following algorithms for connectionist reinforcement learning},
  author={Williams, Ronald J},
  journal={Machine learning},
  volume={8},
  pages={229--256},
  year={1992},
  publisher={Springer}
}

@article{dpo,
  title={Direct preference optimization: Your language model is secretly a reward model},
  author={Rafailov, Rafael and Sharma, Archit and Mitchell, Eric and Manning, Christopher D and Ermon, Stefano and Finn, Chelsea},
  journal={Advances in Neural Information Processing Systems},
  volume={36},
  pages={53728--53741},
  year={2023}
}

@inproceedings{hu2021lora,
  author       = {Edward J. Hu and
                  Yelong Shen and
                  Phillip Wallis and
                  Zeyuan Allen{-}Zhu and
                  Yuanzhi Li and
                  Shean Wang and
                  Lu Wang and
                  Weizhu Chen},
  title        = {LoRA: Low-Rank Adaptation of Large Language Models},
  booktitle    = {The Tenth International Conference on Learning Representations, {ICLR}
                  2022, Virtual Event, April 25-29, 2022},
  year         = {2022},
}

@inproceedings{xie2024show,
  author       = {Jinheng Xie and
                  Weijia Mao and
                  Zechen Bai and
                  David Junhao Zhang and
                  Weihao Wang and
                  Kevin Qinghong Lin and
                  Yuchao Gu and
                  Zhijie Chen and
                  Zhenheng Yang and
                  Mike Zheng Shou},
  title        = {Show-o: One Single Transformer to Unify Multimodal Understanding and
                  Generation},
  booktitle    = {The Thirteenth International Conference on Learning Representations,
                  {ICLR} 2025, Singapore, April 24-28, 2025},
  year         = {2025},
}

@article{wang2024emu3,
  title={Emu3: Next-token prediction is all you need},
  author={Wang, Xinlong and Zhang, Xiaosong and Luo, Zhengxiong and Sun, Quan and Cui, Yufeng and Wang, Jinsheng and Zhang, Fan and Wang, Yueze and Li, Zhen and Yu, Qiying and others},
  journal={arXiv preprint arXiv:2409.18869},
  year={2024}
}

@inproceedings{ma2024janusflow,
  author       = {Yiyang Ma and
                  Xingchao Liu and
                  Xiaokang Chen and
                  Wen Liu and
                  Chengyue Wu and
                  Zhiyu Wu and
                  Zizheng Pan and
                  Zhenda Xie and
                  Haowei Zhang and
                  Xingkai Yu and
                  Liang Zhao and
                  Yisong Wang and
                  Jiaying Liu and
                  Chong Ruan},
  title        = {JanusFlow: Harmonizing Autoregression and Rectified Flow for Unified
                  Multimodal Understanding and Generation},
  booktitle    = {{IEEE/CVF} Conference on Computer Vision and Pattern Recognition,
                  {CVPR} 2025, Nashville, TN, USA, June 11-15, 2025},
  pages        = {7739--7751},
  year         = {2025},
}

@article{chen2025janus,
  title={Janus-pro: Unified multimodal understanding and generation with data and model scaling},
  author={Chen, Xiaokang and Wu, Zhiyu and Liu, Xingchao and Pan, Zizheng and Liu, Wen and Xie, Zhenda and Yu, Xingkai and Ruan, Chong},
  journal={arXiv preprint arXiv:2501.17811},
  year={2025}
}

@inproceedings{xie2025sana,
  author       = {Enze Xie and
                  Junsong Chen and
                  Yuyang Zhao and
                  Jincheng Yu and
                  Ligeng Zhu and
                  Yujun Lin and
                  Zhekai Zhang and
                  Muyang Li and
                  Junyu Chen and
                  Han Cai and
                  Bingchen Liu and
                  Daquan Zhou and
                  Song Han},
  title        = {{SANA} 1.5: Efficient Scaling of Training-Time and Inference-Time
                  Compute in Linear Diffusion Transformer},
  booktitle    = {Forty-second International Conference on Machine Learning, {ICML}
                  2025, Vancouver, BC, Canada, July 13-19, 2025},
  year         = {2025},
}


\appendix

\section{Text-Unit Extraction Prompts}
\label{app:text_unit_prompts}

This appendix reports the task-specific text-unit extraction procedures used by STAR. These text units are used only to locate the corresponding text-token indices and aggregate text-image attention maps; the external image-level reward for each task remains unchanged.

\newtcblisting{PromptBox}{%
  enhanced,
  breakable,
  listing only,
  colback=LakeBlue!3!white,
  colframe=LakeBlue,
  boxrule=0.6pt,
  arc=1mm,
  left=1mm,
  right=1mm,
  top=1mm,
  bottom=1mm,
  before skip=0.8em,
  after skip=0.8em,
  listing options={%
    basicstyle=\ttfamily\footnotesize,
    columns=fullflexible,
    keepspaces=true,
    breaklines=true,
    breakatwhitespace=false,
    showstringspaces=false
  }%
}

\subsection{GenEval Text Units}
\label{app:geneval_text_units}

For GenEval, STAR does not require an LLM for text-unit extraction. Instead, it applies a deterministic focus-span rule: remove a boilerplate image prefix such as ``a photo of'', ``an image of'', or ``a painting of'', and use the remaining phrase as a single text unit. This matches the GenEval prompt format, where the suffix already contains the objects, attributes, counts, and spatial relations to be grounded.

\begin{PromptBox}
GenEval structured-phrase extraction

Input:  c = original GenEval prompt
Rule:   remove prefix matching
        ^(?:a|an|the)?\s*(?:photo|picture|image|painting|drawing)\s+of\s+
Output: K(c) = [remaining focus span]

Examples:
Input:  "a photo of a blue book"
Output: ["a blue book"]

Input:  "a photo of a bus below a cat"
Output: ["a bus below a cat"]

Input:  "a photo of a yellow bus and an orange handbag"
Output: ["a yellow bus and an orange handbag"]
\end{PromptBox}

\subsection{PickScore Text Units}
\label{app:pickscore_text_units}

For PickScore, STAR uses the same deterministic focus-span rule by default. We additionally consider component-level decomposition, in which a vision-language model extracts concise semantic units describing objects, attributes, quantities, and spatial relations. The decomposition prompt is provided below.

\begin{PromptBox}
You are a precise text analyzer. Given an image generation prompt, decompose it into semantic components.

Rules:
- Extract each distinct object with its attributes (e.g., "red cat", "wooden table")
- Extract spatial relationships (e.g., "sitting on", "next to", "above")
- Extract quantities if specified (e.g., "two cups", "three dogs")
- Each component should be a short phrase (1-4 words)
- Return ONLY a JSON list of strings, no explanation

Example:
Input: "a photo of a red cat sitting on a wooden table"
Output: ["red cat", "wooden table", "sitting on"]

Example:
Input: "a photo of two blue cups on a white shelf"
Output: ["two blue cups", "white shelf", "on"]

Example:
Input: "a photo of a black car parked next to a red bicycle"
Output: ["black car", "red bicycle", "parked next to"]

Now decompose:
Input: "{prompt}"
Output:
\end{PromptBox}

\subsection{OCR Text Units}
\label{app:ocr_text_units}

OCR prompts contain exact quoted strings that must be rendered in the image. We therefore extract visually localizable text units in advance and associate them with the corresponding training prompts. These units include the exact rendered text, its carrier or surface, visible text attributes, and other concrete scene elements. When pre-extracted units are unavailable, STAR falls back to the deterministic focus-span rule described above. The extraction prompt is shown below.

\begin{PromptBox}
You are an expert at analyzing OCR/text-to-image prompts. Extract only the important, visually localizable semantic components that should correspond to concrete image regions and cross-attention heatmaps.

Priority for OCR/text-rendering prompts:
1. Exact rendered text: every quoted word/phrase that must appear legibly in the image. Preserve spelling and capitalization.
2. Text carrier/surface: the object or region that contains the text (label, billboard, screen, sign, page, ribbon, poster, display, book cover).
3. Text style/appearance: concrete visible style/color/layout of the rendered text or carrier (red warning label, bold letters, glowing green text).
4. Main objects/subjects: entities that must appear (medicine bottle, robot chest panel, spellbook, astronaut boot print).
5. Concrete scene/background: visually localizable environment (industrial background, Martian surface, neutral background).
6. Composition/viewpoint/style only when visually important and concise (close-up, textbook diagram, realistic photograph).

Include:
- Objects/subjects and their key visible attributes.
- Actions/interactions only when they create a visible region.
- Exact quoted OCR text as a standalone component.
- Text carrier + text content as separate components when both exist.
- Short phrases that can plausibly map to a region in the image.

Exclude:
- Abstract intent/emotion/judgment with no clear visual region: urgency, achievement, accomplishment, vibe, feeling, mood, intrigued, dismissive.
- Generic filler: a photo of, a picture of, an image of, featuring, set against, that reads, clearly reading, prominently displayed.
- Standalone connectors/prepositions: in, on, with, of, that are, a couple of.
- Quality tags: 8k, UHD, highly detailed, masterpiece, trending on Artstation.
- Artist-name modifiers.
- The full prompt itself.

Output format:
Return a JSON array of strings. Each string is one important component, as a short phrase (1-7 words). Prefer 3-7 components per prompt. Keep the original wording when possible. For OCR/text-rendering prompts, include every exact quoted text as one component. Do NOT include the full prompt as a component.

Examples:
Prompt: "A close-up of a medicine bottle with a clear, red warning label that reads \"Take With Food\" prominently displayed, set against a neutral background."
-> ["close-up", "medicine bottle", "clear red warning label", "Take With Food", "neutral background"]

Prompt: "A close-up of a robot's chest panel, with a digital display blinking \"System Override Active\" in red, set against a dimly lit industrial background."
-> ["close-up", "robot's chest panel", "digital display", "System Override Active", "red blinking text", "dimly lit industrial background"]

Batch-format instruction:
Extract the important semantic components from each of the following prompts. Return a JSON array for each prompt, in order. Format: return a JSON array of JSON arrays, one inner array per prompt.

Prompts:
1. {prompt_1}
2. {prompt_2}
...

Return ONLY a JSON array of arrays, e.g.:
[["comp1", "comp2"], ["comp3"], ...]
\end{PromptBox}

\end{document}